\documentclass[runningheads]{llncs}

\usepackage{eccv}

\usepackage{eccvabbrv}
\usepackage{graphicx}
\usepackage{booktabs}
\usepackage[accsupp]{axessibility}

\usepackage{hyperref}
\usepackage{url}
\usepackage{amsmath}
\usepackage{algorithm}
\usepackage{algpseudocode}
\usepackage{amsfonts}
\usepackage{subcaption}
\usepackage{wrapfig}
\usepackage{array}
\usepackage{multirow}
\usepackage{makecell}
\usepackage{pifont}
\usepackage{nicefrac}
\usepackage{microtype}
\usepackage{xcolor}
\usepackage{listings}
\usepackage{amssymb}
\usepackage{adjustbox}
\usepackage{tcolorbox}
\usepackage{tabularx}
\usepackage[utf8]{inputenc}
\usepackage{multicol}
\usepackage{colortbl}
\usepackage{xspace}
\usepackage{orcidlink}

\newcommand{\mname}{\textsl{Compositional Simulation}}

\begin{document}

\title{Building Scalable Real-World Robot Data Generation via Compositional Simulation}

\titlerunning{Compositional Simulation}

\author{Yiran Qin$^{1,6*}$ \quad
Jiahua Ma$^{2*}$ \quad
Li Kang$^{3*}$ \quad
Wenzhan Li$^{2*}$ \\
Yihang Jiao$^{2}$ \quad
Xin Wen$^{2}$ \quad
Xiufeng Song$^{3}$ \quad
Heng Zhou$^{4}$ \quad
Jiwen Yu$^{5}$ \\
Zhenfei Yin$^{6}$ \quad
Xihui Liu$^{5}$ \quad
Philip Torr$^{6}$ \quad
Yilun Du$^{7}$ \quad
Ruimao Zhang$^{2\dagger}$}

\authorrunning{Y.~Qin, J.~Ma, L.~Kang, W.~Li et al.}

\institute{$^{1}$CUHK-Shenzhen \enspace
$^{2}$Sun Yat-sen University \enspace
$^{3}$Shanghai Jiao Tong University \enspace
$^{4}$USTC \\
$^{5}$The University of Hong Kong \enspace
$^{6}$University of Oxford \enspace
$^{7}$Harvard University \\[2pt]
$^{*}$Equal contribution \quad $^{\dagger}$Corresponding author}

\maketitle

\begin{abstract}
Recent advancements in foundational models, such as large language models and world models, have greatly enhanced the capabilities of robotics, enabling robots to autonomously perform complex tasks. However, acquiring large-scale, high-quality training data for robotics remains a challenge, as it often requires substantial manual effort and is limited in its coverage of diverse real-world environments. To address this, we propose a novel hybrid approach called \textbf{\mname{}}, which combines classical simulation and neural simulation to generate accurate action-video pairs while maintaining real-world consistency. Our approach utilizes a closed-loop real-sim-real data augmentation pipeline, leveraging a small amount of real-world data to generate diverse, large-scale training datasets that cover a broader spectrum of real-world scenarios. We train a neural simulator to transform classical simulation videos into real-world representations, improving the accuracy of policy models trained in real-world environments. Through extensive experiments, we demonstrate that our method significantly reduces the sim2real domain gap, resulting in higher success rates in real-world policy model training. Our approach offers a scalable solution for generating robust training data and bridging the gap between simulated and real-world robotics.
\end{abstract}

\section{Introduction}
With the rapid advancements in foundational models, such as large language models~\cite{openai2025gpt5,touvron2023llama,geminiteam2023gemini} and world models~\cite{sora,genie,cosmos}, there has been significant progress in the field of robotics~\cite{du2023learning,yang2023learning}. These innovations have enabled robots to autonomously perform increasingly complex tasks, opening the door to more capable and adaptable robotic systems. Data-driven paradigms have led to impressive results in various domains, yet robotics presents unique challenges compared to fields like language and video modeling. In particular, the need for manually collected video-action pairs poses a significant barrier. Unlike self-supervised learning techniques in other domains, acquiring large-scale data for robotics requires substantial human effort, which is both costly and insufficient for capturing the vast diversity of real-world environments.

While some researchers have addressed this issue by relying on large-scale human data collection, this approach remains expensive and limited in covering the full spectrum of real-world distributions. An alternative method is to leverage simulation to scale data collection~\cite{robocasa2024,mu2024robotwin,qin2025robofactory}, thus reducing the costs associated with real-world data acquisition. Classical simulators, such as MuJoCo~\cite{todorov2012mujoco} and Isaac~\cite{DBLP:conf/nips/MakoviychukWGLS21-isaac}, offer the advantage of generating precise action-video pair data. These simulators use omniscient views, making it easy to generate vast amounts of data with diverse distributions. However, the performance gap between simulated and real-world environments often leads to poor joint training performance when directly transferring simulated data for real-world applications.

\begin{figure}[t]
\begin{center}
\includegraphics[width=1\linewidth]{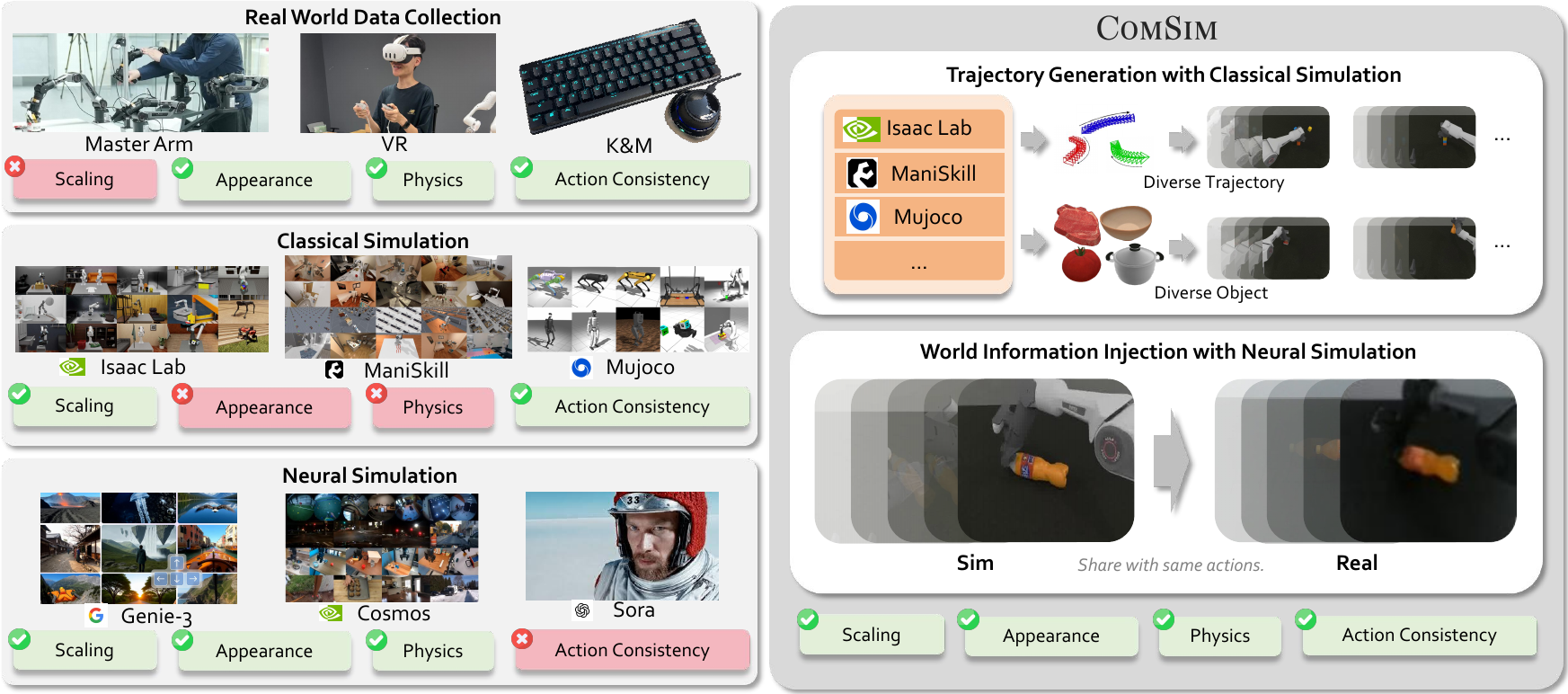}
\end{center}
\vspace{-3mm}
\caption{There are three main sources of real-world robotic data: (1) direct human collection, which yields high-quality samples but cannot scale; (2) classical simulators, which generate large datasets but suffer from appearance and physics gaps to reality; and (3) neural simulators trained on real data, which reduce these gaps but struggle with action-conditioned video generation, leading to weak action–video consistency. We introduce the concept of \mname{}, a flexible and scalable approach that bridges the gap between classical simulation and real-world dynamics via compositional simulation.} 
\label{fig:motivation}
\vspace{-5mm}
\end{figure}

To bridge this gap, neural simulators~\cite{genie,gamefactory,cosmos} based on video generation models~\cite{svd,wan2025wan,opensora} have recently been proposed as a solution. These simulators generate corresponding video data from input trajectories or action signals, producing action-video pairs for training. Although the generated videos appear visually consistent with the real world, issues like hallucination—where videos lack physical consistency and lead to poor action control—undermine the quality of the generated data.

In this work, we introduce the concept of \textbf{\mname{}} shown in Fig~\ref{fig:motivation}, advocating for a hybrid approach that combines the strengths of classical simulation and neural simulation. This approach aims to generate accurate action-video pairs while ensuring that the videos are consistent with real-world dynamics. We propose a closed-loop real-sim-real data augmentation pipeline that utilizes a small amount of real-world data to create training datasets for policy models. These datasets are designed to cover a broader distribution of real-world scenarios. 

The real-sim-real pipeline consists of two key steps. First, we collect a small set of real-world trajectory data and obtain the corresponding videos. In a classical simulation environment, we then replicate the same scenario, replaying the real-world trajectories to generate simulated videos. A neural simulator using Compositional Dynamic Video Generation is subsequently trained to transform the classical simulation videos into real-world videos, ensuring that the actions remain consistent. The second part of our approach involves generating a large and diverse set of action-video pairs through action primitives scheduling within the classical simulator. These pairs are then transformed into real-world representations using the trained neural simulator, facilitating large-scale data augmentation for real-world applications. 
Our main contributions are as follows:
\begin{itemize}
    \item \textit{Concept \& Paradigm.} We introduce the concept of \mname{}, a flexible and scalable approach that bridges the gap between classical simulation and real-world dynamics via compositional simulation.
    \item \textit{Data Pipeline \&  Model.} We propose a real–sim–real data augmentation pipeline that builds a neural simulator with Compositional Dynamic Video Generation ensuring accurate and consistent action–video alignment while simultaneously mitigating the sim2real domain gap.
    \item \textit{Experimental Results.} Extensive experiments demonstrate that \mname{} substantially enhances the policy models by simultaneously increasing task success rates and achieving strong generalization across both spatial layouts and object variations.
\end{itemize}

\section{Related Work}
\label{related_word}


\subsection{Robotic Simulation}
General-purpose simulators such as Isaac Lab~\cite{DBLP:conf/nips/MakoviychukWGLS21-isaac} and MuJoCo~\cite{todorov2012mujoco} provide GPU-parallel capabilities and ready-to-use environments for reinforcement and imitation learning. Frameworks including RoboCasa~\cite{robocasa2024}, Habitat~\cite{szot2021habitat}, AI2-THOR~\cite{ai2thor}, OmniGibson~\cite{li2022behavior}, and RoboFactory~\cite{qin2025robofactory} offer logically structured APIs that standardize action interfaces, enable systematic domain randomization, and facilitate large-scale dataset generation. ManiSkill3 builds upon the open-source SAPIEN simulator~\cite{Xiang_2020_SAPIEN} for GPU-parallel simulation, emphasizing clean action semantics and deterministic stepping to enforce logical action consistency. Despite these advancements, current simulators still fall short of real-world fidelity, with persistent gaps in appearance statistics, sensor characteristics, and contact physics—limitations that often hinder robust policy transfer to real-world dynamics.

\subsection{Data Augmentation for Robotic Manipulation} Recent works have explored leveraging synthetic data to enhance policies in robotic manipulation. Real-to-sim-to-real pipelines, such as URDFormer~\cite{chen2024urdformer}, RialTo~\cite{torne2024reconciling} and~\cite{dai2024automated}, demonstrate effective policy transfer by constructing simulation environments from real-world data. Diffusion models have also been employed for data augmentation in robotics, with ~\cite{yu2023scaling} using inpainting to generate diverse object appearances and backgrounds. Other synthetic demonstration generation methods, including MimicGen~\cite{mandlekar2023mimicgen} and DemoGen~\cite{xue2025demogen}, adapt action trajectories to novel object configurations. While these approaches advance the field, our work specifically targets systematic evaluation of model generalization under controlled scene variations.

\subsection{Robot Learning in Manipulation}
Specialized policy architectures~\cite{chi2023diffusion,ke20243d,adaptdiffuser,liang2024skilldiffuser,dexhanddiff,wang2024rise,wen2025dexvla,Ze2024DP3} often excel on narrowly defined tasks yet struggle to carry over to new robot embodiments. In contrast, foundation models trained on million-scale, multi-robot corpora exhibit strong zero-shot transfer: RT-1~\cite{brohan2022rt-1} unifies vision, language, and action in a single transformer for real-time kitchen manipulation; RT-2~\cite{brohan2023rt-2} jointly finetunes large vision–language models on web and robot data to support semantic planning and object reasoning; diffusion-based RDT-1B~\cite{liu2024rdt1b} and $\pi$\cite{black2024pi_0} learn diverse bimanual dynamics from over a million episodes. Vision–language–action systems such as OpenVLA\cite{openvla} and CogACT~\cite{li2024cogact}, together with adaptations like Octo~\cite{octo_2023}, LAPA~\cite{lapa}, and OpenVLA-OFT~\cite{openvla_oft}, further demonstrate efficient finetuning across robots and sensing modalities. Collectively, these results point to a data-driven bottleneck: robust cross-task and cross-embodiment generalization hinges on large, diverse, and high-fidelity datasets that faithfully capture real-world appearance, sensing, and physics.

\subsection{World Simulator for Robotic Manipulation} 
Scalable robot learning~\cite{bjorck2025gr00t,brohan2022rt,zitkovich2023rt,cheang2024gr,lynch2023interactive} depends on abundant, realistic data, yet collecting real-world trajectories via human demonstrations is slow and labor-intensive, limiting broad access. Generative video models~\cite{agarwal2025cosmos,wu2023unleashing} offer a cost-effective way to synthesize policy training data. UniPi~\cite{du2023learning} and AVDC~\cite{ko2023learning} cast robot planning as text-to-video generation (AVDC further estimates inverse dynamics with a pretrained flow network); UniSim~\cite{yang2023learning} learns a unified real-world simulator across text and control inputs; RoboDreamer~\cite{zhou2024robodreamer} targets compositional generalization via text parsing; and IRASim~\cite{zhu2024irasim} performs trajectory-conditioned video generation but focuses on arm motion only. In this work, our world simulator turns action-consistent simulation trajectories into high-fidelity, real-style data.

\section{Compositional World Simulation}

\subsection{Problem Formulation}

In the context of robotic manipulation, collecting real-world data is often a challenging and resource-intensive task. Traditional methods leverage classical simulators~\cite{todorov2012mujoco,DBLP:conf/nips/MakoviychukWGLS21-isaac,maniskill2} to train online reinforcement learning policies~\cite{DBLP:journals/corr/SchulmanWDRK17-ppo}. These simulators generate large amounts of trajectory data by simulating various robot behaviors. Another approach~\cite{mu2024robotwin,qin2025robofactory} utilizes pre-designed primitive functions, called via large language models~(LLMs), to generate extensive trajectory data, thereby aiming to cover as much of the decision space as possible. These trajectories are commonly used for pre-training or joint training with real-world data.



Despite the large volume of video-action pairs generated, the disparity between the distributions of simulated and real-world data creates significant challenges. Let $\mathcal{D}_{\text{sim}} = \{(v_i, a_i)\}_{i=1}^N$ represent the dataset of video-action pairs collected from a classical simulator, where $v_i$ denotes the video frame and $a_i$ the corresponding action. Similarly, let $\mathcal{D}_{\text{real}} = \{(v'_j, a'_j)\}_{j=1}^M$ represent the real-world dataset, where $v'_j$ and $a'_j$ are the video and action pairs from the real world. Directly training policies on the combined simulated and real data often fails to improve performance or generalization, as the domain gap between simulation and reality exacerbates this issue, leading to degraded policy performance in real-world settings. This gap is particularly evident in appearance and physics, where simulated data cannot fully capture the complexities of the real world.

An alternative method involves using video generation models as neural world simulators. These models generate data that is intended to be as close as possible to real-world distributions. However, video generation models suffer from inherent issues, such as hallucinations, 3D scene consistency, and inaccurate action control. As a result, the generated actions and corresponding videos do not align perfectly, making this data unsuitable for policy training.

To address these issues, we propose a compositional simulation approach. In this approach, we first collect a large number of trajectories in a classical simulator, $\mathcal{D}_{\text{sim}}$. These trajectories are then transformed into video representations using a pre-trained neural simulator $\mathcal{N}$, which maps the simulated data into the real-world distribution. Crucially, this process ensures that the generated data maintains action alignment with the original simulated trajectories. Formally, we aim to build a neural simulation function $\mathcal{N}(\cdot)$, such that:

\begin{equation}
\mathcal{N}(\mathcal{D}_{\text{sim}}) \approx \mathcal{D}_{\text{real}}
\end{equation}

This neural simulation function $\mathcal{N}(\cdot)$ maps the simulated video-action pairs to a distribution that is as close as possible to the real-world data, ensuring that the generated action $a_i$ aligns with the original action $a'_j$, where $a_i \approx a'_j$. Additionally, the consistency of the 3D scene and the video quality must be maintained, addressing the inherent challenges in video generation models. Thus, we transform the simulated data $\mathcal{D}_{\text{sim}}$ to approximate the real-world distribution $\mathcal{D}_{\text{real}}$, while ensuring that the generated actions and videos are consistent with real-world expectations. By applying this compositional simulation approach, we can effectively utilize the large-scale data generated in simulators and adapt it to real-world environments, thereby mitigating the challenges posed by domain gaps in robotic manipulation tasks.

\begin{figure*}[t]
  \centering
  \includegraphics[width=1.0\textwidth]{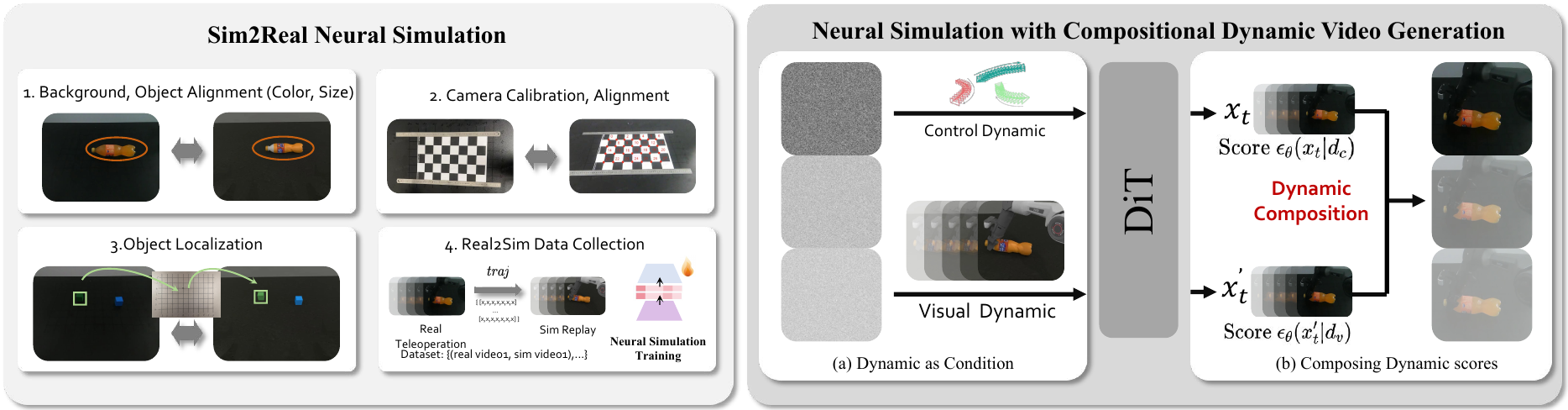}
  \caption{(Left) Alignment between real-world and simulation: trajectories collected in the real world are replayed in simulation to generate paired video data for training the sim-to-real neural simulator. (Right) A DiT can be used to estimate scores conditioned on different dynamics, including Control Dynamics~(actions) and Visual Dynamics~(simulated observations). These scores can be composed during sampling to enable Dynamic Guidance methods.}
  \label{fig:wrap_half}
\end{figure*}


\subsection{Real2Sim Data Collection}
\label{sec:sim2real_neural_simulation}

To train the Sim2Real neural simulation that maps videos to real-world distributions while maintaining the correct actions, we need to construct a dataset composed of tuples $(\mathcal{V}_{\text{sim}}, \mathcal{V}_{\text{real}}, \mathcal{A})$, where $\mathcal{V}_{\text{sim}}$ and $\mathcal{V}_{\text{real}}$ represent the results of executing the same action in the classical simulator and the real world, respectively. In other words, $\mathcal{V}_{\text{sim}}$ and $\mathcal{V}_{\text{real}}$ share the same action sequence.

To build such a dataset, we need to create a simulation data collection platform that aligns strictly with the real-world data collection platform. As shown in Fig.~\ref{fig:wrap_half}~(Left), to establish this digital twin simulation environment, we performed alignment at three levels:

\textbf{Background and Object Alignment:} We first aligned the background and objects in the simulation, including their colors and sizes. The desktop and background colors in the classical simulator were aligned with those of the real-world data collection platform. Additionally, we applied a digital twin approach to assets to ensure visual consistency and real-world scale matching.

\textbf{Camera Calibration and Alignment:} We then calibrated and aligned the cameras to ensure that the camera parameters and poses in the real world were consistent with those in the classical simulation.

\textbf{Object Position Alignment:} During task initialization, we localized the objects in the real-world scene and strictly transferred their position information into the classical simulator.

\subsection{Sim2Real Compositional Dynamic Video Generation}

To ensure precise action control and its seamless transfer to the real world, we train the neural simulator using Compositional Dynamic Video Generation.  We leverage Control Dynamics (actions) and Visual Dynamics (simulated observations) to generate video sequences simulating real-world scenarios. As shown in Fig.~\ref{fig:wrap_half}~(Right), the DiT is conditioned on these dynamics, allowing it to estimate future states based on current actions and visual feedback. The scores, conditioned on both Control and Visual Dynamics, are composited during the sampling process to generate a comprehensive representation of the evolving system state. This compositional approach enables the DiT to adaptively combine the influences of action and observation, enhancing generalization for complex, dynamic tasks. The composited scores guide the sampling process, ensuring that the video generation aligns closely with the task-specific dynamics, thus enabling Dynamic Guidance methods that steer the simulation towards accurate and robust video generation.

Specifically, we collect data from the real-world simulation platform in Sec.~\ref{sec:sim2real_neural_simulation} to obtain the pair $(\mathcal{V}_{\text{real}}, \mathcal{A})$. These action data are then replayed in the corresponding classical simulation environment to generate the tuple $(\mathcal{V}_{\text{sim}}, \mathcal{V}_{\text{real}}, \mathcal{A})$. We collected data for 10 tasks, resulting in 200 data pairs for training.
To optimize the neural simulator for Sim2Real data generation, we aim to minimize the discrepancy between the simulated and real-world videos, while maintaining the correct action alignment. Since the actions in both $\mathcal{V}_{\text{sim}}$ and $\mathcal{V}_{\text{real}}$ are already aligned, we focus solely on optimizing the video consistency. The optimization objective is formulated as:


\begin{equation}
\mathcal{L}_{\text{sim2real}} = \mathcal{L}_{\text{video}}(f_{\mathcal{N}}(\mathcal{V}_{\text{sim}}, \mathcal{A}, \theta), \mathcal{V}_{\text{real}})
\end{equation}

Where $\mathcal{L}_{\text{video}}$ measures the difference between the generated simulated video and the real-world video, and $\theta$ represents the neural simulator's parameters. By minimizing this loss, the neural simulator learns to generate videos that closely match the real-world distribution, while preserving the correct action alignment.

\begin{figure}[t]
\begin{center}
\includegraphics[width=1\linewidth]{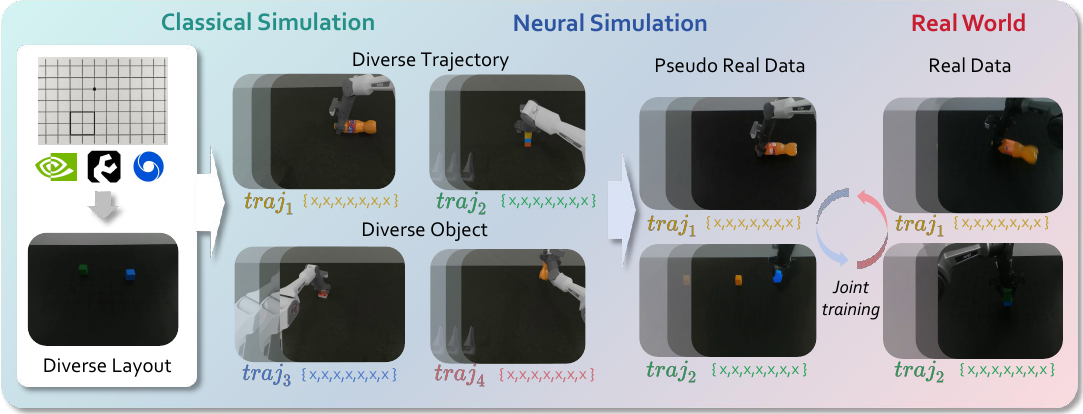}
\end{center}
\vspace{-3mm}
\caption{\textbf{Real World Deployment with \mname{}.} Large volumes of $(\mathcal{V}_{\text{sim}}, \mathcal{A})$ pairs are collected from the classical simulator and transformed into corresponding $(\mathcal{V}_{\text{real}}, \mathcal{A})$ pairs, referred to as \textit{Pseudo Real Data}. These data, together with a small amount of real-world data, are used to train policies with improved success rates and generalization.} 
\label{fig:pipeline}
\vspace{-5mm}
\end{figure}

\subsection{Data Generation with Rule-Based Simulation}
\label{sim_data}
To further scale up the data collection pipeline, we employ RoboTwin~\cite{chen2025robotwin}, a SAPIEN-based~\cite{Xiang_2020_SAPIEN} dual-arm manipulation simulation environment. It provides a rich library of digital assets and supports diverse trajectory distributions, making it well-suited for synthesizing large-scale visuomotor datasets. By systematically varying environmental conditions, object initialization states, and agent actions, we generate an extensive set of trajectories and corresponding videos that cover a broad spectrum of real-world task scenarios.

Specifically, we define a comprehensive set of interaction rules, referred to as action primitives, governing how agents and objects interact within the simulation. These primitives serve as the atomic building blocks of complex behaviors, capturing low-level manipulations (e.g., grasp, push, align) as well as higher-order skills (e.g., stack blocks). We curate a suite of RoboTwin tasks and adapt them to support richer interaction patterns and object configurations, enabling a broader spectrum of physical reasoning scenarios. To automate the generation of complex behaviors, we employ GPT-5~\cite{openai2025gpt5} to synthesize executable code composed of these action primitives, while integrating compositional constraints~\cite{qin2025robofactory} to ensure semantic correctness and physical feasibility. The action primitives encompass a variety of object types and interaction modalities, enabling diverse scenario generation. For each task, we construct a rich collection of trajectories ${\tau_s}$ spanning the action space, and carefully tune the primitive-based generation process to achieve comprehensive coverage. This allows us to traverse the global distribution of agent behaviors in the simulation, including different object initializations and heterogeneous object categories.

The resulting dataset comprises temporally synchronized camera observations, corresponding action and state sequences. These elements are strictly aligned at the behavioral level, ensuring that every visual frame is paired with its underlying control command. Although the trajectories and interactions in simulation are faithful to their intended semantics, the rendered appearance of the videos still differs from real-world imagery due to discrepancies in lighting, textures, and sensor noise. To bridge this domain gap, we pass the simulated observation streams $v_s$ through a neural simulator $\mathcal{N}$, which refines their visual characteristics while preserving the original dynamics and action consistency.

\begin{algorithm}
\caption{Real World Deployment with Compositional Simulation}
\begin{algorithmic}[1]
\State \textbf{Input:} 
\State \quad Classical simulation data $(\mathcal{V}_{\text{sim}}, \mathcal{A})$, Real-world data $(\mathcal{V}_{\text{real}}, \mathcal{A})$
\State \textbf{Functions:} 
\State \quad Neural Simulator $\mathcal{N}$, Video Transformation Function $f_{\mathcal{N}}$
\State \textbf{Hyperparameters:} 
\State \quad Real-World Data Ratio $\alpha$
\State Initialize $D_{\text{sim}} \leftarrow \{\mathcal{V}_{\text{sim}}, \mathcal{A}\}$, $D_{\text{real}} \leftarrow \{\mathcal{V}_{\text{real}}, \mathcal{A}\}$ \Comment{Initialize datasets}
\State $P_{\text{pseudo}} \leftarrow \{\}$ \Comment{Initialize Pseudo Real Data set}
\For{each $(\mathcal{V}_{\text{sim}}, \mathcal{A}) \in D_{\text{sim}}$}
    \State $P_{\text{pseudo}} \leftarrow P_{\text{pseudo}} \cup f_{\mathcal{N}}(\mathcal{V}_{\text{sim}}, \mathcal{A})$ \Comment{Transform simulation data to Pseudo Real Data}
\EndFor
\State $D_{\text{combined}} \leftarrow \alpha \cdot D_{\text{real}} + (1 - \alpha) \cdot P_{\text{pseudo}}$ \Comment{Combine Pseudo Real Data with Real Data}
\State Train policy $\pi_{\text{robot}}$ using $D_{\text{combined}}$ \Comment{Train robot policy using combined data}
\State \textbf{Return:} Trained robot policy $\pi_{\text{robot}}$
\end{algorithmic}
\end{algorithm}

\subsection{Real World Deploy with Compositional Simulation}

As shown in Fig.~\ref{fig:pipeline}, after training the neural simulator, we proceeded with the process outlined in Sec.~\ref{sim_data} to collect a large number of $(\mathcal{V}_{\text{sim}}, \mathcal{A})$ pairs from the classical simulation. These data are then fed into the neural simulator, which transforms them into corresponding $(\mathcal{V}_{\text{real}}, \mathcal{A})$ pairs. We refer to these transformed data as \textit{Pseudo Real Data}. Compared to the data produced by classical simulators, these Pseudo Real Data exhibit representations that are much closer to real-world data, with a reduced domain gap.

By using these Pseudo Real Data, which cover a broader distribution of scenarios, in conjunction with a small amount of real-world data collected from the actual environment, we can jointly train a robot policy. This approach significantly improves the performance and generalization capability of the policy. The specific experimental results are presented in the Sec.~\ref{exp_realworld}.

\section{Experiments}
\label{sec:exp}

\subsection{Sim2Real Transfer via Neural Simulation}
\textbf{Baselines.}
\label{subsec:baselines}
To validate the effectiveness of our proposed Neural Simulation in recovering real-world data distributions from simulation, we consider the following set of diverse comparative approaches:
1) Classical Simulation(Sim), denoting the canonical raw simulation pipeline without neural-driven refinement;
2) Baseline, a video-to-video generation model built on Stable Diffusion 1.5~\cite{rombach2022high} with temporal continuity post-processing~\cite{yang2024fresco}; 
3) Zero-Shot, referring to the backbone model deployed without any sim-to-real fine-tuning;
4) Ours-CD, a variant of our proposed Neural Simulation framework, equipped with conditional generation capability guided solely by control dynamics for pseudo-realistic content synthesis;
5) Ours-VD, an alternative variant of our method, featuring conditional generation functionality driven exclusively by visual dynamics;
and 6) Ours-Full, the full instantiation of our pipeline, with joint control-visual dynamics conditional generation capability.
Comparisons across all approaches naturally form an ablation study to empirically evaluate our variants’ efficacy in bridging the sim-to-real gap. For fair comparison, our evaluation protocol is: all models take an identical simulation video and sim-to-real instruction as input, to generate the corresponding pseudo-realistic video.

\textbf{Quantitative Results.}
For quantitative evaluation, we employ a set of widely used perceptual and structural similarity metrics (PSNR, SSIM, CLIP Score~\cite{hessel2021clipscore}, LPIPS~\cite{zhang2018unreasonable}), alongside distributional measures (FID~\cite{heusel2017gans}, FVD~\cite{unterthiner2018towards}), to assess both the visual fidelity of the generated videos with respect to real-world videos and their temporal coherence throughout the frame sequence.
We conduct a comprehensive evaluation of all competing methods on a test suite comprising 8 distinct tasks: 
\textit{Shake Bottle}, \textit{Stack Blocks Two}, \textit{Move Playing-Card Away}, \textit{Move Playing-Card Away (Cluttered)}, \textit{Move Playing-Card Away (Colored Background)}, \textit{Place Mouse Pad (Cluttered)}, \textit{Place Mouse Pad (Colored Background)}, and \textit{Handover Bottle (Cluttered)}. 
The quantitative comparison results are summarized in Tab.~\ref{tab:sim2real_ablation}. 
First, conventional UNet-based diffusion model baseline exhibits subpar overall performance, while the base video generation model without sim-to-real domain adaptation conversely degrades the photorealism of generated results. Further, we observe that our Neural Simulation variant guided solely by control dynamics (Ours-CD) delivers negligible effectiveness, revealing that visual dynamics information plays a dominant role in the joint conditional generation process. Aligned with this, our full pipeline (Ours-Full) achieves top performance across all metrics. This demonstrates that joint control and visual dynamics guidance enables high perceptual realism video synthesis, validating its effectiveness in bridging the intrinsic sim-to-real gap.

\begin{table}[t]
  \centering
  \footnotesize
  \vspace{-3mm}
  \setlength{\tabcolsep}{8pt}  
  \caption{Comparison of the realism quality of generated videos across different methods.}
  \label{tab:sim2real_ablation}
  \resizebox{0.9\textwidth}{!}{%
    \begin{tabular}{ccccccc}
    \toprule
    & \textbf{PSNR} $\uparrow$ & \textbf{SSIM} $\uparrow$ & 
    \textbf{CLIP Score} $\uparrow$ & \textbf{LPIPS} $\downarrow$ & \textbf{FID} $\downarrow$ & \textbf{FVD} $\downarrow$ \\
    \midrule
    \textbf{Sim} & 16.443 & 0.4342 & 0.7564 & 0.3629 & 187.40 & 61.048 \\
    \textbf{Baseline} & 16.849 & 0.5129 & 0.7526 & 0.3494 & 254.59 & 50.369 \\
    \textbf{Zero-Shot} & 13.093 & 0.5487 & 0.7308 & 0.4756 & 219.74 & 163.83 \\
    \textbf{Ours-CD} & 8.4640 & 0.1486 & 0.7216 & 0.8130 & 434.44 & 239.13 \\
    \textbf{Ours-VD} & 18.153 & 0.5916 & 0.7884 & 0.2813 & 153.12 & 22.311 \\
    \textbf{Ours-Full} & \textbf{19.577} &  \textbf{0.6484} &  \textbf{0.8102} &  \textbf{0.2647} &  \textbf{147.90} &  \textbf{15.765} \\
    \bottomrule
    \end{tabular}
  }
  \label{tab:evaluation_metrics}%
\end{table}

\begin{figure}[t]
    \vspace{-2mm}
    \centering
    \includegraphics[width=1.0\linewidth]{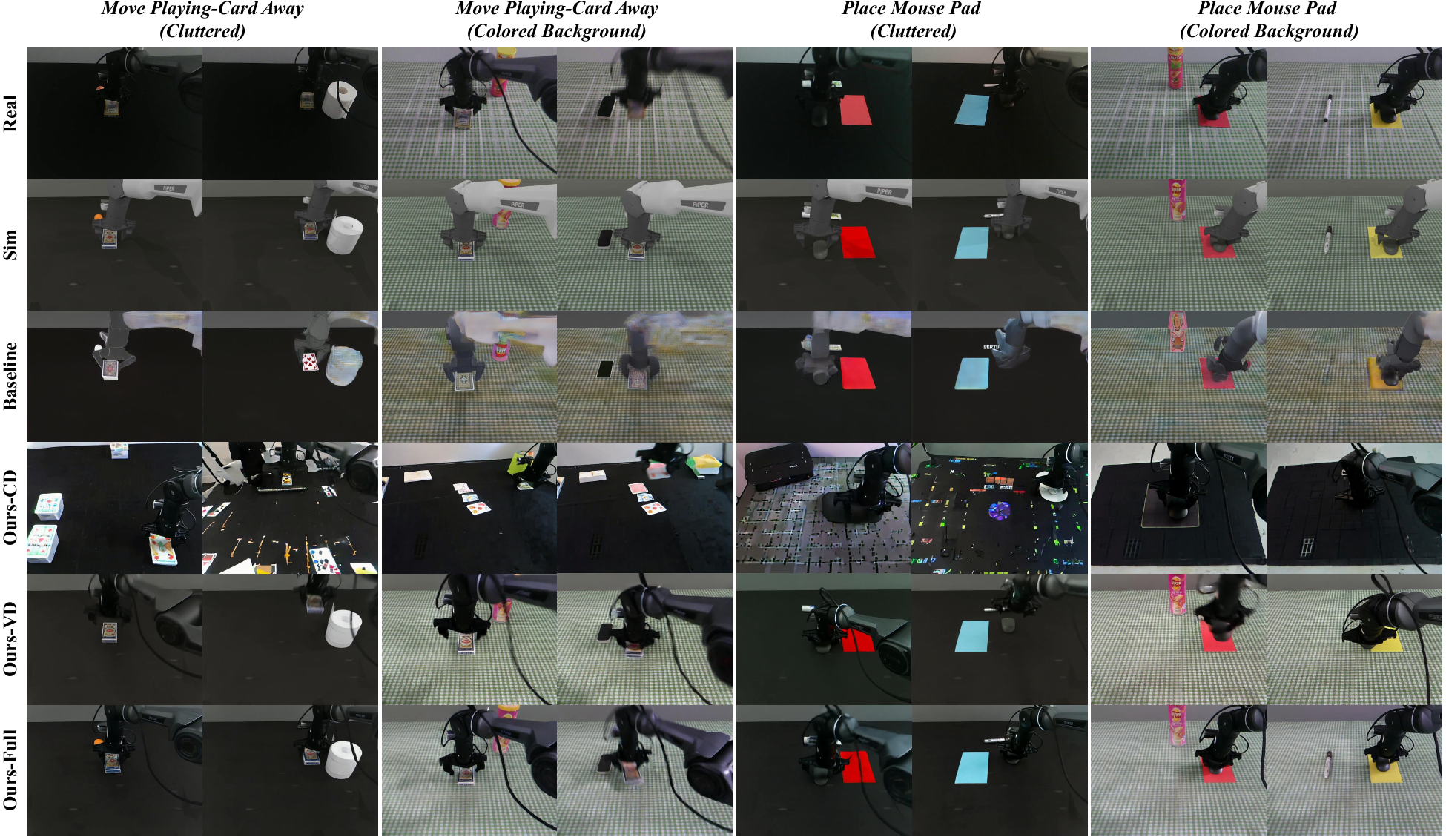}
    \caption{
    Visual comparison of generated results across four different tasks. 
    }
    \vspace{-5mm}
    \label{fig:sim2real_res}
\end{figure}

\textbf{Qualitative Results.}
We conduct a visual comparison across four representative and generalization-demonstrative tasks, as shown in Fig.~\ref{fig:sim2real_res}.
These tasks are explicitly designed to cover generalization variations across two core dimensions: object attributes (including category and color) and desktop background configurations. For visualization, we select frames that either fully present the complete robotic gripper or capture the key interaction moments of each task, while strictly ensuring that the generated results across different tasks and all compared methods are sampled from the identical frame index.
Note that the simulated objects exhibit visual discrepancies from their real-world counterparts in appearance. In particular, the robotic gripper appears dark black in reality, as opposed to white-gray in the simulator.

As visualized in Fig.~\ref{fig:sim2real_res}, while conventional diffusion baselines exhibit reasonable scene understanding capability, they fail to effectively preserve critical visual semantic information during generation, resulting in severe hallucination artifacts in their outputs. In comparison, our control dynamics-only variant (Ours-CD) accurately reproduces the target manipulation actions, yet lacks the scene semantic constraints provided by visual dynamics information. This leads to the viewpoint, object instances, and desktop background being purely generated from the model’s inherent prior knowledge, resulting in unsatisfactory overall performance. Conversely, our visual dynamics-only variant (Ours-VD) contributes the majority of our framework’s sim-to-real transfer capability, while still exhibiting non-negligible flaws in two core dimensions: action accuracy (including the opening/closing timing of the robotic gripper and overall task completion quality) and physical consistency (such as motion blur matching the robotic arm’s movement). Finally, our full pipeline (Ours-Full) not only achieves photorealistic visual generation of the agent and scene, but also leverages motion guidance from control dynamics to enable accurate reproduction of real-world manipulation actions, ultimately realizing precise sim-to-real alignment across both visual perception and action execution.

\subsection{Real world Execution with Compositional World Simulation}
\label{exp_realworld}

\label{sec:real_world_execution_with_compositional_world_simulation}

\begin{table*}[ht]
  \centering
  \footnotesize
  \vspace{-3mm}
  \setlength{\tabcolsep}{4pt}  
 \caption{Quantitative Results of Our Real-world Experiments. The compared methods include DP trained under six data-mixture regimes: \textit{10 Real}, \textit{20 Real}, \textit{200 Sim Pretrain + 10 Real}, \textit{10 Real + 200 Sim}, \textit{10 Real + 200 Pseudo Real}, and \textit{200 Pseudo Real}. ``OOD'' abbreviates out-of-distribution.}
  \resizebox{0.9\textwidth}{!}{%
    \begin{tabular}{>{\centering\arraybackslash}m{3.5cm}*{4}{>{\centering\arraybackslash}m{2.5cm}}}
    \hline
       \textbf{\makecell[c]{Real World\\Task}}  & \textbf{Spatial Distribution} & \textbf{\makecell[c]{10 Real}} & \textbf{\makecell[c]{20 Real}} & \textbf{\makecell[c]{200 Sim Pretrain \\ + 10 Real}} \\
    \hline
    
    \multirow{2}{*}{\textit{Shake Bottle}} &
        In Domain & 9/30 & 17/30 & 12/30 \\
        & OOD  & 0/30 & 1/30 & 0/30 \\
    \hline
    
    \multirow{2}{*}{\textit{Stack Blocks Two}} & 
        In Domain & 5/30 & 13/30 & 8/30 \\
        & OOD & 0/30 & 0/30 & 0/30 \\
    \hline

    \multirow{2}{*}{\textit{Move Playing-Card Away}} & 
        In Domain & 12/30 & 24/30 & 15/30 \\
        & OOD & 0/30 & 3/30 & 2/30 \\
    \hline
    
    \multirow{2}{*}{\makecell{\textit{Move Playing-Card Away} \\ \textit{(Cluttered)}}} & 
        In Domain & 7/30 & 18/30 & 11/30 \\
        & OOD & 0/30 & 1/30 & 1/30 \\
    \hline

    \multirow{2}{*}{\makecell{\textit{Move Playing-Card Away} \\ \textit{(Colored Background)}}} & 
        In Domain & 10/30 & 20/30 & 6/30 \\
        & OOD & 0/30 & 2/30 & 1/30 \\
    \hline

    \multirow{2}{*}{\makecell{\textit{Place Mouse Pad} \\ \textit{(Cluttered)}}} & 
        In Domain & 4/30 & 15/30 & 6/30 \\
        & OOD & 0/30 & 0/30 & 1/30 \\
    \hline

    \multirow{2}{*}{\makecell{\textit{Place Mouse Pad} \\ \textit{(Colored Background)}}} & 
        In Domain & 7/30 & 18/30 & 8/30 \\
        & OOD & 0/30 & 1/30 & 1/30 \\
    \hline

    \multirow{2}{*}{\makecell{\textit{Handover Bottle} \\ \textit{(Cluttered)}}} & 
        In Domain & 3/30 & 8/30 & 1/30 \\
        & OOD & 0/30 & 0/30 & 0/30 \\
    \hline
    
    \bottomrule
       \textbf{\makecell[c]{Real World\\Task}}  & \textbf{Spatial Distribution} & \textbf{\makecell[c]{10 Real \\ + 200 Sim
        }} & \textbf{\makecell[c]{10 Real + 200 \\ Pseudo Real}} & \textbf{\makecell[c]{200 Pseudo Real\\(Zero Shot)}} \\
    \hline
    
    \multirow{2}{*}{\textit{Shake Bottle}} & 
        In Domain & 6/30 & \textbf{28/30} & 10/30 \\
        & OOD & 0/30 & \textbf{12/30} & 5/30 \\
    \hline
    
    \multirow{2}{*}{\textit{Stack Blocks Two}} & 
        In Domain & 2/30 & \textbf{15/30} & 7/30 \\
        & OOD & 0/30 & \textbf{6/30} & 3/30 \\
    \hline

    \multirow{2}{*}{\textit{Move Playing-Card Away}} & 
        In Domain & 6/30 & \textbf{29/30} & 18/30 \\
        & OOD & 1/30 & \textbf{17/30} & 9/30 \\
    \hline
    
    \multirow{2}{*}{\makecell{\textit{Move Playing-Card Away} \\ \textit{(Cluttered)}}} & 
        In Domain & 3/30 & \textbf{25/30} & 15/30 \\
        & OOD & 1/30 & \textbf{16/30} & 8/30 \\
    \hline

    \multirow{2}{*}{\makecell{\textit{Move Playing-Card Away} \\ \textit{(Colored Background)}}} & 
        In Domain & 8/30 & \textbf{23/30} & 17/30 \\
        & OOD & 2/30 & \textbf{17/30} & 10/30 \\
    \hline

    \multirow{2}{*}{\makecell{\textit{Place Mouse Pad} \\ \textit{(Cluttered)}}} & 
        In Domain & 7/30 & \textbf{19/30} & 8/30 \\
        & OOD & 1/30 & \textbf{10/30} & 5/30 \\
    \hline

     \multirow{2}{*}{\makecell{\textit{Place Mouse Pad} \\ \textit{(Colored Background)}}} & 
        In Domain & 5/30 & \textbf{22/30} & 12/30 \\
        & OOD & 1/30 & \textbf{14/30} & 9/30 \\
    \hline

     \multirow{2}{*}{\makecell{\textit{Handover Bottle} \\ \textit{(Cluttered)}}} & 
        In Domain & 1/30 & \textbf{13/30} & 11/30 \\
        & OOD & 0/30 & \textbf{5/30} & 3/30 \\
    \hline
    
    \end{tabular}
  }
  \label{tab:real-world-results}%
\end{table*}

\textbf{Baselines.} To rigorously quantify the benefit of our proposed compositional world simulation pipeline under an extremely limited real-world demonstration budget, we trained six instances of Diffusion Policy (DP)~\cite{chi2023diffusion} according to the following data-mixture regimes: 
1) 10 Real: learning solely from 10 real-world demonstrations. 
2) 20 Real: doubling the real-world budget to 20 demonstrations to isolate the gain of additional real-world data. 
3) 200 Sim Pretrain + 10 Real: pre-training on 200 RoboTwin-simulated demonstrations followed by fine-tuning on the same 10 real-world demonstrations used in Regime 1. 
4) 10 Real + 200 Sim: jointly training on the 200 RoboTwin-simulated and 10 real-world demonstrations from scratch. All demonstrations used here are same as Regime 3. 
5) 10 Real + 200 Pseudo-Real: 
jointly training on the 200 pseudo-real demonstrations, which were generated by our compositional world simulation pipeline previously, and 10 real demonstrations used in Regime 1 from scratch.
6) 200 Pseudo-Real (Zero-Shot): zero-shot training exclusively on the 200 pseudo-real demonstrations used in Regime 5, establishing an upper-bound on the performance achievable by DP without any real-world supervision.

\begin{figure}[t]
\begin{center}
\includegraphics[width=1\linewidth]{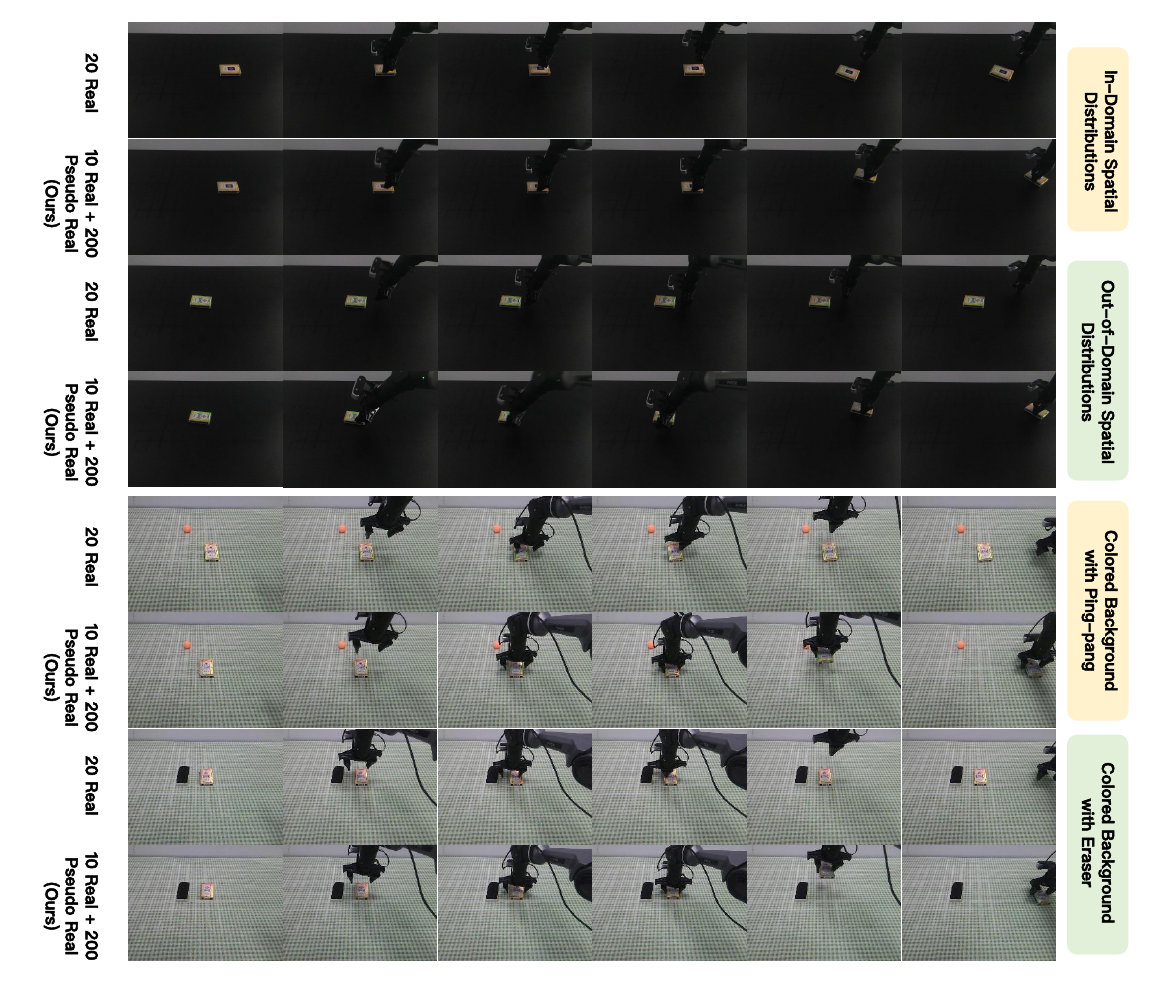}
\end{center}
\vspace{-3mm}
\caption{Visualization of DP performance on \textit{Move Playing-Card Away}. Top two rows: objects lie initially within the region predefined in collected real-world demonstrations (in-domain spatial distribution). Middle two rows: initial positions are outside the region (out-of-domain spatial distribution). Bottom four rows: introduce a colored background with varying levels of object clustering. Policies shown are trained under \textit{20 Real} and \textit{10 Real + 200 Pseudo Real}, respectively.}
\label{fig:generalization_spatial}
\vspace{-4mm}
\end{figure}

\textbf{Quantitative Results.}
As shown in Tab.~\ref{tab:real-world-results}, DP performs poorly when only 10 real-world demonstrations are available (cf. 10 Real), and its success rate improves steadily as more real-world data are provided (cf. 20 Real), underscoring the importance of sufficient real-world experience. Simulated data from the traditional simulator (such as RoboTwin) also helps, yet the benefit is capped by the visual–physical gap between real-world and simulated environments (cf. 200 Sim Pretrain + 10 Real and 10 Real + 200 Sim). In contrast, the Pseudo-Real demonstrations generated by our compositional world simulation pipeline narrow this gap and yield a substantial increase in task success rate (cf. 10 Real + 200 Pseudo Real), and even delivering non-trivial performance in the complete absence of real data (200 Pseudo Real).

\subsection{Generalization}

To further validate the fidelity of our pipeline in reproducing real-world scenarios, we conducted an ablation study on the ability of DP to generalize to new spatial layouts and new objects. All DP evaluated here are identical to those introduced in Sec.~\ref{sec:real_world_execution_with_compositional_world_simulation}.

\textbf{Generalization to Novel Spatial Distributions.}  
It is necessary to note the initialized position of every object in the collected real-world demonstrations was confined to a predefined limited region (see Sup.~8.2). At inference, we relocated the same objects to previously unseen regions and recorded the success rates of all DPs. Tab.~\ref{tab:real-world-results} and Fig.~\ref{fig:generalization_spatial} shows that DPs trained solely on real data exhibit almost zero generalization to the new regions; the spatial diversity present in traditional RoboTwin simulations is likewise rendered ineffective by the sim-to-real gap, yielding no measurable improvement. In contrast, the pseudo-real demonstrations generated by our compositional world simulation pipeline consistently lift performance across the relocated configurations, confirming that the synthesized scenes faithfully reproduce the spatial statistics of the real-world environments.

\begin{figure}[h]
\begin{center}
\includegraphics[width=1\linewidth]{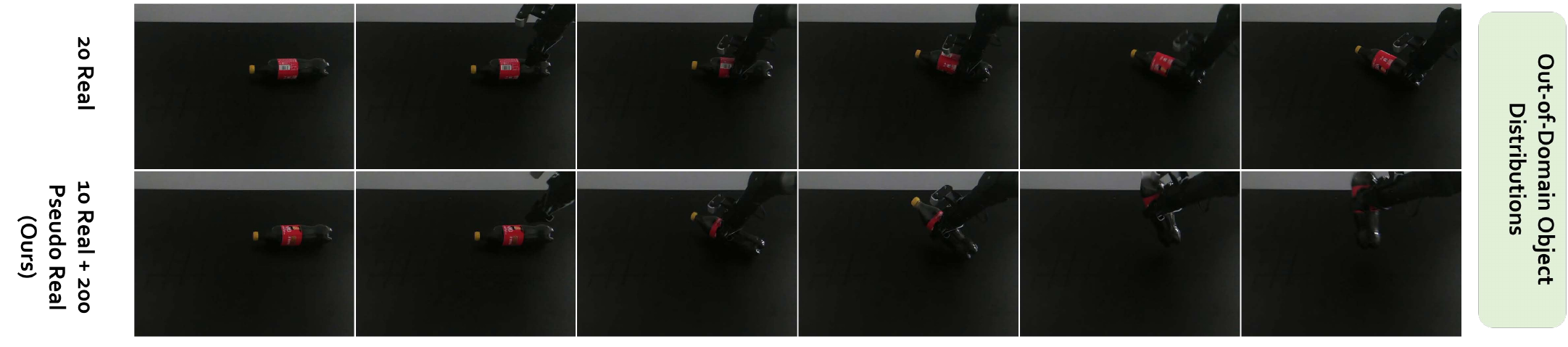}
\end{center}
\vspace{-5.5mm}
\caption{Generalization visualization of DP on \textit{Shake Bottle} under out-of-domain object distributions. Top: policy trained with \textit{20 Real}. Bottom: policy trained with \textit{10 Real + 200 Pseudo Real}.}
\label{fig:generalization_objects}
\vspace{-5mm}
\end{figure}

\textbf{Generalization to New Objects.} We evaluate shape- and color-level generalization by substituting new objects at inference time. Concretely, in the real-world demonstrations we employ a Fanta bottle and a blue playing card, and at inference time these are replaced by other bottles (i.e. Coca-Cola, Sprite and Nongfu Spring Oriental Leaf Tea) and a red playing card, respectively.
As shown in Table~\ref{tab:real-world-results-generalization} and Fig.~\ref{fig:generalization_objects}, simulated demonstrations collected from RoboTwin bring no improvement in the generalization to new objects, whereas the Pseudo-Real demonstrations generated by our compositional world simulation pipeline yield a clear boost in success rate.
This demonstrates that our method preserves real-world properties and supports transfer to unseen objects.

\begin{table*}[t]
  \centering
  \footnotesize
  \setlength{\tabcolsep}{4pt}  
  \caption{Quantitative evaluation of DP new-object generalization across six data mixtures.}
  \resizebox{0.82\textwidth}{!}{
    \begin{tabular}{@{}
      >{\centering\arraybackslash}m{4cm}  
      *{3}{>{\centering\arraybackslash}m{3cm}}
      @{}}
      
    \hline
       \textbf{\makecell[c]{Real World Task}} & \textbf{\makecell[c]{10 Real}} & \textbf{\makecell[c]{20 Real}} & \textbf{\makecell[c]{200 Sim Pretrain \\ + 10 Real}} \\
    \hline
    
   \textit{Shake Bottle} &
        0/30  & 0/30 &  0/30 \\

    \textit{Move Playing-Card Away} &
       1/30 & 2/30 &  1/30 \\
    \bottomrule
       \textbf{\makecell[c]{Real World Task}} & \textbf{\makecell[c]{10 Real \\ + 200 Sim
        }} & \textbf{\makecell[c]{10 Real + 200 \\ Pseudo Real}} & \textbf{\makecell[c]{200 Pseudo Real\\(Zero Shot)}} \\
    \hline
    
    \textit{Shake Bottle} & 
        0/30 & \textbf{15/30} & 9/30  \\
    
    \textit{Move Playing-Card Away} &
       0/30 & \textbf{21/30} &  11/30 \\
    \hline
    \end{tabular}
  }
  \label{tab:real-world-results-generalization}
  \vspace{-3mm}
\end{table*}

\section{Conclusion}
We presented \mname{}, a hybrid framework that integrates classical and neural simulation through a real–sim–real pipeline to generate accurate and consistent action–video pairs. Our approach leverages limited real-world data to create large-scale, diverse training datasets, substantially narrowing the sim2real domain gap. Experiments show that \mname{} improves real-world policy success rates and enables stronger generalization across tasks, spaces, and objects. This work offers a scalable path toward robust data generation for embodied intelligence and opens avenues for extending to richer modalities and broader robotic embodiments.

\textbf{Limitation and Future Work.} Our experiments focus on tabletop manipulation, though the framework could be extended to more complex embodiments such as mobile manipulation with wheeled robot. Future work may try to use of unpaired data to enhance capability and generalization, which would further advance compositional simulation.



\bibliographystyle{splncs04}


\section*{Appendix}
\section{Use of LLMs}
This paper was written by the authors without any generative contribution from large language models (LLMs). LLMs were employed solely for language polishing and grammatical refinement; no scientific content, technical claims, or novel interpretations were produced or altered by these tools.

\section{Task Details}
\label{sec:app_task_details}

To facilitate assets alignment between the real-world and simulated settings, we select five representative tasks in RoboTwin~\cite{chen2025robotwin}—\textit{Shake Bottle}, \textit{Move Playing Card Away}, \textit{Stack Blocks Two}, \textit{Place Mouse Pad} and \textit{Handover bottle}—to evaluate our compositional world simulation framework. Their respective success criteria are defined as follows.
\begin{itemize}
\item \textit{Stack Blocks Two} utilizes two colored blocks—green and yellow. This task is designed primarily to assess the model’s ability to generalize to novel spatial configurations. Success is achieved when the robot first places the green block at the designated position and subsequently stacks the yellow block precisely on top of it.

\item \textit{Shake Bottle} involves four beverages—Fanta, Coca-Cola, Sprite, and Nongfu Spring Oriental Leaf Tea. Among these beverages, Fanta is employed to collect real-world demonstrations, while the remaining three serve as new objects for an ablation study on model generalization. The task is deemed successful if the robot grasps the bottle from the desktop, lifts it to a predefined height, and performs a shaking motion.

\item \textit{Move Playing-Card Away} employs two types of playing cards that differ in color—blue and red. Following the same protocol as \textit{Shake Bottle}, the blue playing card is used to construct the real-world training dataset, whereas the red playing card serves as an unseen object for evaluating model generalization. The task is considered successful once the robot grasps the designated card and transports it completely away from the central region of the desktop. 

\item \textit{Place Mouse Pad} requires the robot to grasp a mouse and place it precisely onto a designated colored pad. Success is achieved when the mouse is fully positioned on the target pad.

\item \textit{Handover bottle} is specifically designed to evaluate the model's robustness when handling dynamic objects. The task involves two robotic arms working in coordination. It is considered successful when the robot successfully grasps the bottle and hands it over to another robot in a seamless manner.
\end{itemize}

To systematically assess the model capabilities beyond the standard task settings, we introduce two types of scene variations for a subset of tasks:
\begin{itemize}
\item \textit{Cluttered Scene:} For \textit{Move Playing-Card Away}, \textit{Place Mouse on Pad}, and \textit{Handover Bottle}, we introduce task-irrelevant distractors (e.g., office supplies, miscellaneous objects) placed randomly on the desktop. This protocol evaluates the model's ability to distinguish the target object from distractors in complex environments.

\item \textit{Background Shift:} For \textit{Move Playing-Card Away} and \textit{Place Mouse on Pad}, we alter the environmental background (e.g., changing desktop texture and color). This protocol assesses the model's robustness to contextual variations and ensures it does not overfit to specific background features observed during training.
\end{itemize}

\section{Training Details}
\label{app:training_details}
\subsection{Neural Simulator Training Details}
As mentioned in 
the main text,
our Neural Simulator builds upon Stable Diffusion 1.5~\cite{rombach2022high}, a state-of-the-art latent text-to-image diffusion model capable of generating high-fidelity visual content from textual prompts. We provide a fixed sim-to-real instruction as its text input, namely: \textit{``Change the image style from the image style of the simulated environment to the image style captured by a DSLR camera.''}. Next, we pair the initial simulation data produced by our Classical Simulator with corresponding real-world data to form simulation–real data pairs. The base model is then fine-tuned on these pairs by minimizing the diffusion model’s denoising loss. Finally, an online inference strategy FRESCO~\cite{yang2024fresco} is applied to the fine-tuned model to generate the final high-quality pseudo-realistic videos.

All experiments are conducted on one NVIDIA H200 GPU. During fine-tuning, the video data is first converted into image sequences at 10 FPS and organized into a training set, with $1 / 5$ of the data randomly sampled as a validation set. The model is trained for 30 epochs with a batch size of 8 and a gradient accumulation step of 4, taking approximately 10 hours. We employ the \texttt{torch.optim.AdamW} optimizer with a learning rate of $5.0 \times 10^{-5}$ and a linear warm-up ratio of 0.01. For the loss function, the diffusion model’s denoising loss is empirically weighted by 1.0, the perceptual loss (measuring feature-level differences between generated and real images) is weighted by 0.2, and the pixel-wise loss (computing the RGB mean squared error between generated and real images) is weighted by 0.1. During inference, we follow the default parameter settings of FRESCO, except that the minimum key-frame sampling interval is set to 3, to ensure smoother video generation.

\subsection{DP Training Details}
\subsubsection{Demonstrations} 
\label{app:demonstrations}
\begin{figure}[t]
\begin{center}
\includegraphics[width=1\linewidth]{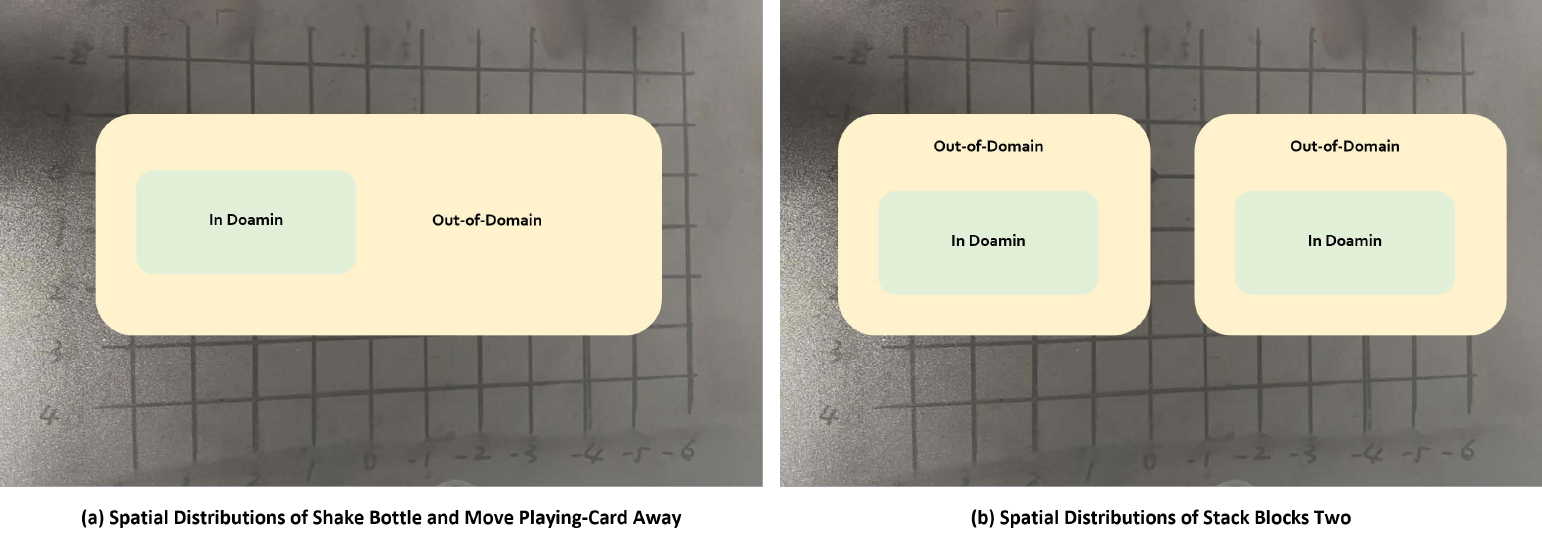}
\end{center}
\caption{Definition of in-domain and out-of-domain spatial distributions in different tasks. Both terms refer exclusively to the initial position of objects before being manipulated. Positions are labeled in-domain if and only if they appear in the collected real-world demonstrations; all others are deemed out-of-domain.}
\label{fig:spatial-distribution}
\end{figure}

\textbf{Real-World Demonstrations} were meticulously collected via human teleoperation using a pair of PiPER Teach Pendants (see Sup.~\ref{app:evaluation_platform}). For each task, we recorded only 20 trajectories, all confined to the in-domain spatial and object distribution. Concretely, demonstrations for \textit{Shake Bottle} and \textit{Move Playing-Card Away} were acquired exclusively with the Fanta bottle and the blue playing card starting within the in-domain region illustrated as Fig.~\ref{fig:spatial-distribution}(a), respectively. And for \textit{Stack Blocks Two}, the green and yellow blocks were always placed in the left and right in-domain zones at the beginning of demonstration collection (Fig.~\ref{fig:spatial-distribution}(b)). Finally, we randomly selected 10 out of these 20 demonstrations to construct the data-mixture regime \textit{10 Real}, and used all 20 to construct the regime \textit{20 Real}.

\textbf{Simulated Demonstrations} were generated in the traditional simulator RoboTwin, and we collected 200 trajectories for each task. In contrast to the real-world ones, these simulated demonstrations deliberately incorporated out-of-domain spatial arrangements and objects. Specifically, for \textit{Shake Bottle} we used not only the Fanta bottle but also Coca-Cola, Sprite and Nongfu Spring Oriental Leaf Tea, while for \textit{Move Playing-Card Away} we included the red playing card in addition to the blue one. Besides, all objects might be placed in out-of-domain regions when data collection started. All 200 simulated demonstrations were employed to construct the data-mixture regimes \textit{10 Real + 200 Sim} and \textit{200 Sim Pre-train + 10 Real}.

\textbf{Pseudo-Real Demonstrations} were produced by our compositional world-simulation framework under the same out-of-domain spatial and object settings employed for Simulated Demonstrations. The full set of 200 pseudo-real trajectories was used to establish the data-mixture regimes \textit{10 Real + 200 Pseudo-Real} and \textit{200 Pseudo-Real}.

\subsubsection{Training Settings} 
We use Diffusion Policy (DP)~\cite{chi2023diffusion}, a generative method based on imitation learning. 
We employ a CNN-based Diffusion Policy as the backbone of our visuomotor model. The prediction horizon is set to 8, with 3 observation steps and 6 action steps.
For data loading, we use a batch size of 256. The optimizer is \texttt{torch.optim.AdamW} with a learning rate of $1.0 \times 10^{-4}$, betas in $[0.95, 0.999]$, and $\epsilon = 1.0 \times 10^{-8}$.
A learning-rate warmup is applied for the first 500 steps, followed by 300 training epochs for all benchmark tasks.

Each policy is trained independently on a single NVIDIA H200 GPU for 300 epochs.
As a reference, using a dataset of roughly 200 demonstration episodes (average length $\approx 300$), training a single policy for 300 epochs takes about 20 hours.

\section{Evaluation Details}
\label{app:evaluation_details} 

\subsection{Platform} 
\label{app:evaluation_platform} 
\begin{figure}[t]
\begin{center}
\includegraphics[width=1\linewidth]{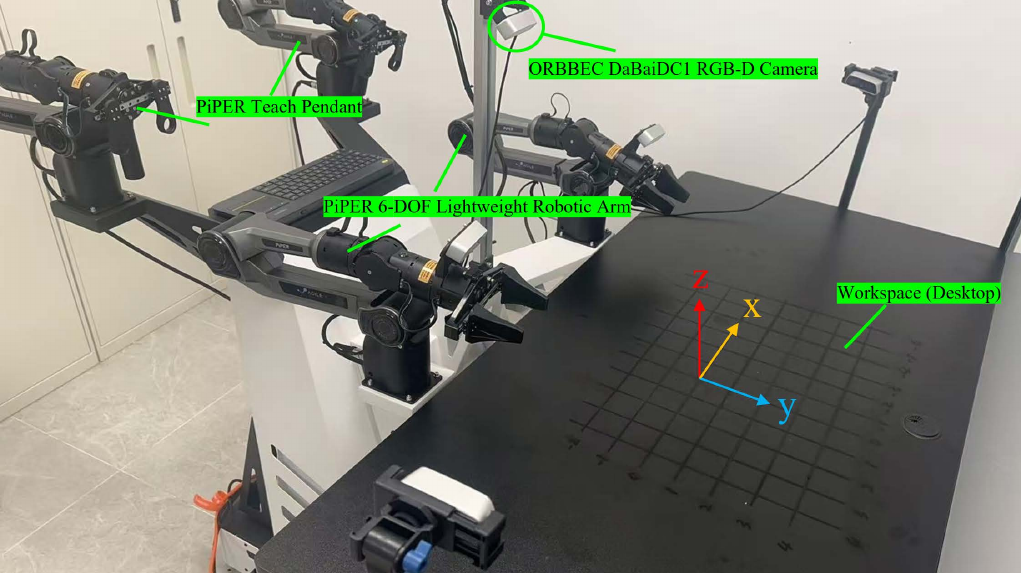}
\end{center}
\caption{Real-world evaluation platform.} 
\label{fig:workspace}
\end{figure}

As depicted in Fig.~\ref{fig:workspace}, the real-world evaluation were performed with two ORBBEC PiPER 6-DOF Lightweight Robotic Arms, each equipped with a two-finger gripper (maximum opening 70 mm, gripping force 40 N). A fixed, top-down ORBBEC DaBaiDC1 RGB-D camera provided a global RGB view of the workspace, defined as the central area of a black tabletop. In addition, a pair of PiPER Teach Pendants enabled teleoperation of the arms, allowing efficient collection of real-world demonstrations. All hardware units were connected to a workstation housing an NVIDIA GeForce RTX 4090 GPU, which stored the captured observations, performed model inference, issued control commands, and drove the robotic arms in real time.

\subsection{Evaluation Settings} 
For each ablation dimension—in-domain spatial/object configurations, out-of-domain spatial layouts, and out-of-domain objects—we independently constructed a fixed set of 30 diverse real-world trials. Every policy trained under a different data-mixture regime was evaluated on the corresponding 30-trial split, guaranteeing that all comparisons within a distribution are performed on an identical test bed. Task-success criteria are provided in Sup.~\ref{sec:app_task_details}.

\section{Sim2Real Neural Simulation Details}
\label{app:sim2real_neural_simulation_details}
As stated in the main text,
we enforce strict alignment between the real-world and simulated environments—encompassing background and object appearance, camera intrinsics/extrinsics, and object positions—to enable effective Sim2Real neural simulation. To prevent any learning-induced errors from propagating into the subsequent training of the neural simulator, we adopt a purely rule-based alignment pipeline rather than a data-driven one. Concretely, we first parameterize the relevant attributes of the real-world scene and then transfer the estimated parameters to configure the simulated environment accordingly. 

\subsection{Background and Object Alignment}

\textbf{Background Alignment} mainly parameterizes both the visual appearance of the desktop and the laboratory walls. Using the fixed RGB-D camera described in Sec.~\ref{app:evaluation_platform}, we first capture images of the table surface and the wall regions. A digital color-picker is then applied to the acquired images to extract representative RGB values.

\textbf{Regular-Object Alignment} covers primitives such as blocks, spheres, and cylinders whose geometry can be described by a small set of metric dimensions. For these instances, we first measure their principal axes (length, width, height, diameter, etc.) with calipers. Their Appearance are parameterized by acquiring an orthographic RGB patch of the object’s most representative face and extracting the median albedo via a color-picker tool—no additional texture map is required, yielding a compact, error-tolerant representation.

\textbf{Special-Object Alignment.} Owing to RoboTwin’s one-to-one digital twins of real-world assets—including Finda, Fanta, Coca-Cola, Sprite, Nongfu Spring Oriental Leaf Tea bottles and the playing cards—we can directly pair every physical item with its pre-modeled, dimension- and texture-matched counterpart. This eliminates the need for on-the-fly scanning or manual modeling: each real-world bottle or card is simply mapped to its pre-registered URDF/FBX model, guaranteeing sub-millimetre geometric agreement and pixel-level texture consistency between reality and simulation.

\subsection{Camera Calibration and Alignment}

Camera parameterization focuses on retrieving its intrinsic and extrinsic.
The intrinsics can be known directly from its technical documentation; hence, the following section details only the extrinsic-calibration pipeline employed in our setup.

To ensure consistency with RoboTwin, we establish a real-world coordinate system as illustrated in Fig.~\ref{fig:workspace}. Within this coordinate system, we first place a calibration checkerboard and obtain the 3D coordinates of its corner points. Subsequently, we capture images using the mounted camera and extract the 2D pixel coordinates of those checkerboard corners via corner detection. The code implementing this procedure is listed below.
\begin{lstlisting}[
    language=Python,
    basicstyle=\ttfamily\scriptsize,
    numbers=left,
    numberstyle=\tiny\color{gray},
    frame=single,
    breaklines=true,
    keywordstyle=\color{blue}\bfseries,
    commentstyle=\color{gray}\itshape,
    stringstyle=\color{red},
    label={lst:evaluation_code},
]
import cv2
import numpy as np

# --------------------------
PATTERN_SIZE = (7, 4)          # Number of checkerboard corners along rows and columns
IMG_PATH     = 'chess.jpg'     # Path of the captured image
# --------------------------

img  = cv2.imread(IMG_PATH)
if img is None:
    raise FileNotFoundError(IMG_PATH)
gray = cv2.cvtColor(img, cv2.COLOR_BGR2GRAY)

# 1. Checkerboard Detection
ret, corners = cv2.findChessboardCorners(
    gray, PATTERN_SIZE,
    cv2.CALIB_CB_ADAPTIVE_THRESH + cv2.CALIB_CB_FAST_CHECK + cv2.CALIB_CB_NORMALIZE_IMAGE)
if not ret:
    raise RuntimeError('Checkerboard detection failed!')

corners = cv2.cornerSubPix(
    gray, corners, (11, 11), (-1, -1),
    criteria=(cv2.TERM_CRITERIA_EPS + cv2.TERM_CRITERIA_MAX_ITER, 30, 0.001))
pts2d = corners.reshape(-1, 2).tolist()

# 2. Mouse Callback
def on_mouse(event, x, y, flags, param):
    global pts2d, img_show
    if event == cv2.EVENT_LBUTTONDOWN:
        idx = min(range(len(pts2d)),
                  key=lambda i: (pts2d[i][0] - x) ** 2 + (pts2d[i][1] - y) ** 2)
        if (pts2d[idx][0] - x) ** 2 + (pts2d[idx][1] - y) ** 2 < 400:
            del pts2d[idx]
            redraw()

def redraw():
    global img_show
    img_show = img.copy()
    for idx, (u, v) in enumerate(pts2d):
        u, v = int(u), int(v)
        cv2.circle(img_show, (u, v), 5, (0, 0, 255), -1)
        cv2.putText(img_show, str(idx), (u + 10, v - 10),
                    cv2.FONT_HERSHEY_SIMPLEX, 0.5, (0, 0, 255), 1, cv2.LINE_AA)
    cv2.imshow('interactive', img_show)

# 3. Main Loop
cv2.namedWindow('interactive', cv2.WINDOW_NORMAL)
cv2.setMouseCallback('interactive', on_mouse)
redraw()

print('Usage:')
print('  Left-click on a corner -> toggle delete/undelete')
print('  Press q -> save current pts2d.txt and exit')
print('  Press r -> restore all originally detected corners')
print('  Close the window (x) to quit without saving')

while True:
    if cv2.getWindowProperty('interactive', cv2.WND_PROP_VISIBLE) < 1:
        break

    key = cv2.waitKey(30) & 0xFF  # 30 ms timeout to prevent freezing
    if key == ord('q'):
        np.savetxt('pts2d.txt', np.array(pts2d), fmt='%.6f')
        print('Saved pts2d.txt with', len(pts2d), 'points.')
        break
    elif key == ord('r'):
        pts2d = corners.reshape(-1, 2).tolist()
        redraw()

cv2.destroyAllWindows()
\end{lstlisting}

With the obtained 3D-to-2D correspondences, the camera’s extrinsic parameters—position and orientation—can be recovered by solving a Perspective-n-Point (PnP) problem. The implementation is given below.
\begin{lstlisting}[
    language=Python,
    basicstyle=\ttfamily\scriptsize,
    numbers=left,
    numberstyle=\tiny\color{gray},
    frame=single,
    breaklines=true,
    keywordstyle=\color{blue}\bfseries,
    commentstyle=\color{gray}\itshape,
    stringstyle=\color{red},
    label={lst:extrinsic_code},
]
import numpy as np
import cv2

# 1. Given Known Intrinsics
K = np.array([
    [488.8112487792969, 0.0, 317.05938720703125],
    [0.0, 488.8112487792969, 217.4825439453125],
    [0.0, 0.0, 1.0]
], dtype=np.float64)

dist = np.zeros((4, 1))  # Distortion Coefficients; set to zero if no distortion.

# 2. Input
pts3d = np.loadtxt('pts3d.txt')   # Nx3, World Coordinates
pts2d = np.loadtxt('pts2d.txt')   # Nx2, Pixel Coordinates

assert pts3d.shape[0] == pts2d.shape[0], 'Mismatch in point count!'

# 3. Estimating Extrinsic
ok, rvec, tvec = cv2.solvePnP(
    pts3d.astype(np.float64),
    pts2d.astype(np.float64),
    K, dist, flags=cv2.SOLVEPNP_ITERATIVE
)
if not ok:
    raise RuntimeError('Solve PnP failed! Please verify that the point correspondences are correct.')

# Rotation Vector -> Rotation Matrix
R, _ = cv2.Rodrigues(rvec)

R_w2c = R 
t_w2c = tvec.ravel()
R_c2w = R_w2c.T
t_c2w = -R_c2w @ t_w2c   # 3x1

position = t_c2w.tolist()

forward = R_c2w[:, 2]
left    = -R_c2w[:, 0]
up      = -R_c2w[:, 1]

print("static_camera_list:")
print("  - name: head_camera")
print("    type: D435")
print(f"    position: [{position[0]:.3f}, {position[1]:.3f}, {position[2]:.3f}]")
print(f"    forward:  [{forward[0]:.3f}, {forward[1]:.3f}, {forward[2]:.3f}]")
print(f"    left:     [{left[0]:.3f}, {left[1]:.3f}, {left[2]:.3f}]")
\end{lstlisting}

\subsection{Object Position Alignment}
We emphasize that the real-world frame depicted in Fig.~\ref{fig:workspace} coincides exactly with the world frame employed in RoboTwin. Under this frame, the desktop surface is tessellated into a uniform 5 cm × 5 cm lattice. Object placement is thereby reduced to aligning the object’s center of mass with a lattice node; orientation is selected from a prescribed, rule-based catalogue—namely, axis-aligned poses or rotations of 30°, 45°, and 60° about the x- or y-axis.
While this discrete parameterization is admittedly naive, it routinely delivers positional errors below one centimetre and angular errors below one degree. A data-driven, continuous 6-DoF alignment module will be investigated in future work to supersede this manual gridding scheme.

\section{More Result Visualization}

\subsection{Visualization of Generalization on New Objects}
\begin{figure}[H]
\begin{center}
\includegraphics[width=1\linewidth]{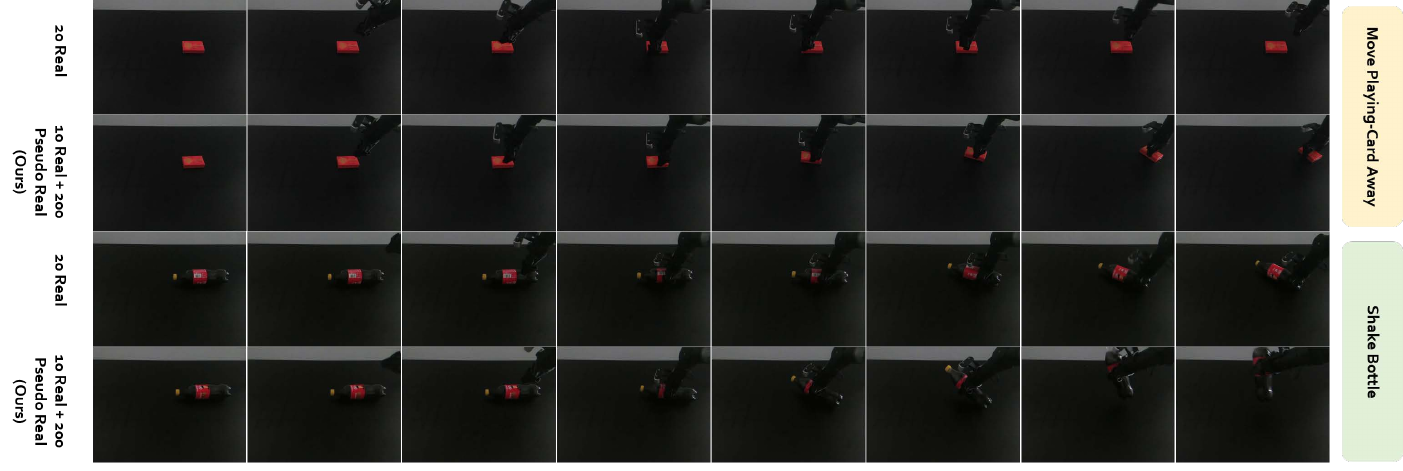}
\end{center}
\caption{Generalization visualization of DP on new objects. The top two rows are corresponds to \textit{Move Playing-Card Away}, and the bottom two rows are corresponds to \textit{Shake Bottle}. respectively.}
\label{fig:app_generalization_objects}
\vspace{-40mm}
\end{figure}

To highlight the enhanced generalization of DP enabled by our compositional world simulation pipeline, we provide trajectory visualizations of the \textit{Move Playing-Card Away} task 
with additional examples in Fig.~\ref{fig:app_generalization_objects}.

\subsection{Visualization of Real2Sim Alignment}
\begin{figure}[H]
\begin{center}
\includegraphics[width=1\linewidth]{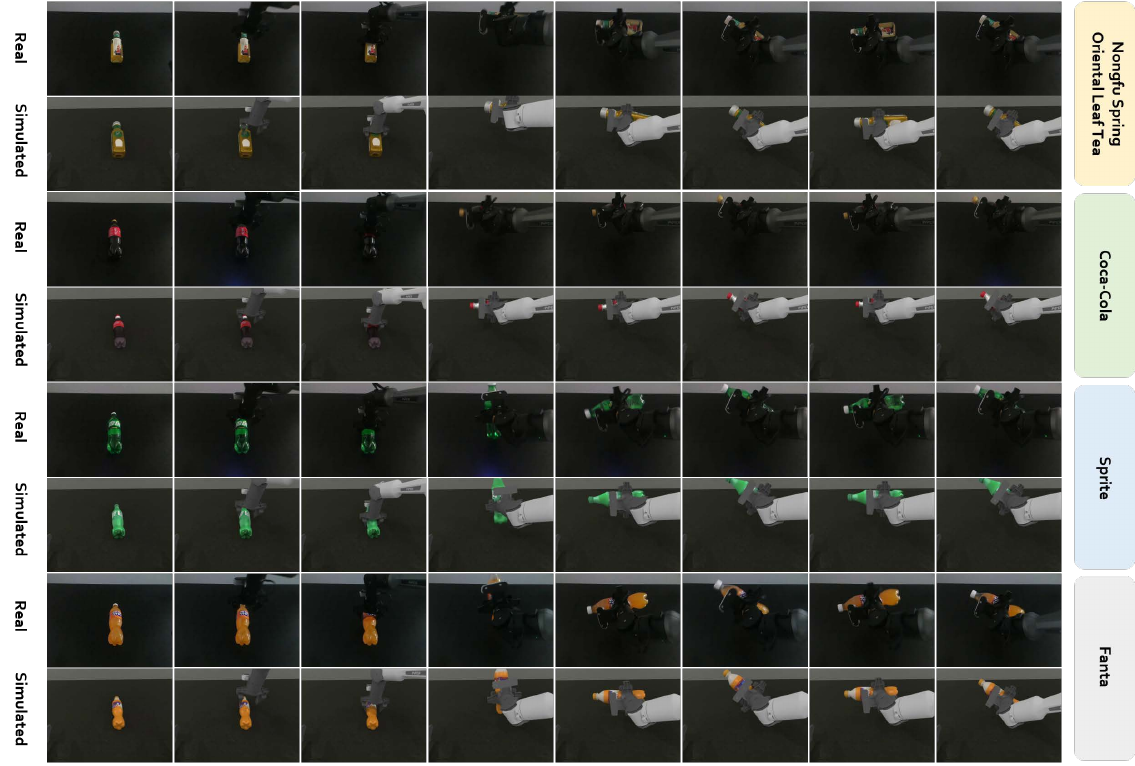}
\end{center}
\vspace{10mm}
\caption{Real2Sim alignment on \textit{Move Playing-Card Away}. From top to bottom: Nongfu Spring Oriental Leaf Tea, Coca-Cola, Sprite, and Fanta.}
\label{fig:app_visual_shake_bottle}
\vspace{10mm}
\end{figure}

\begin{figure}[H]
\begin{center}
\includegraphics[width=1\linewidth]{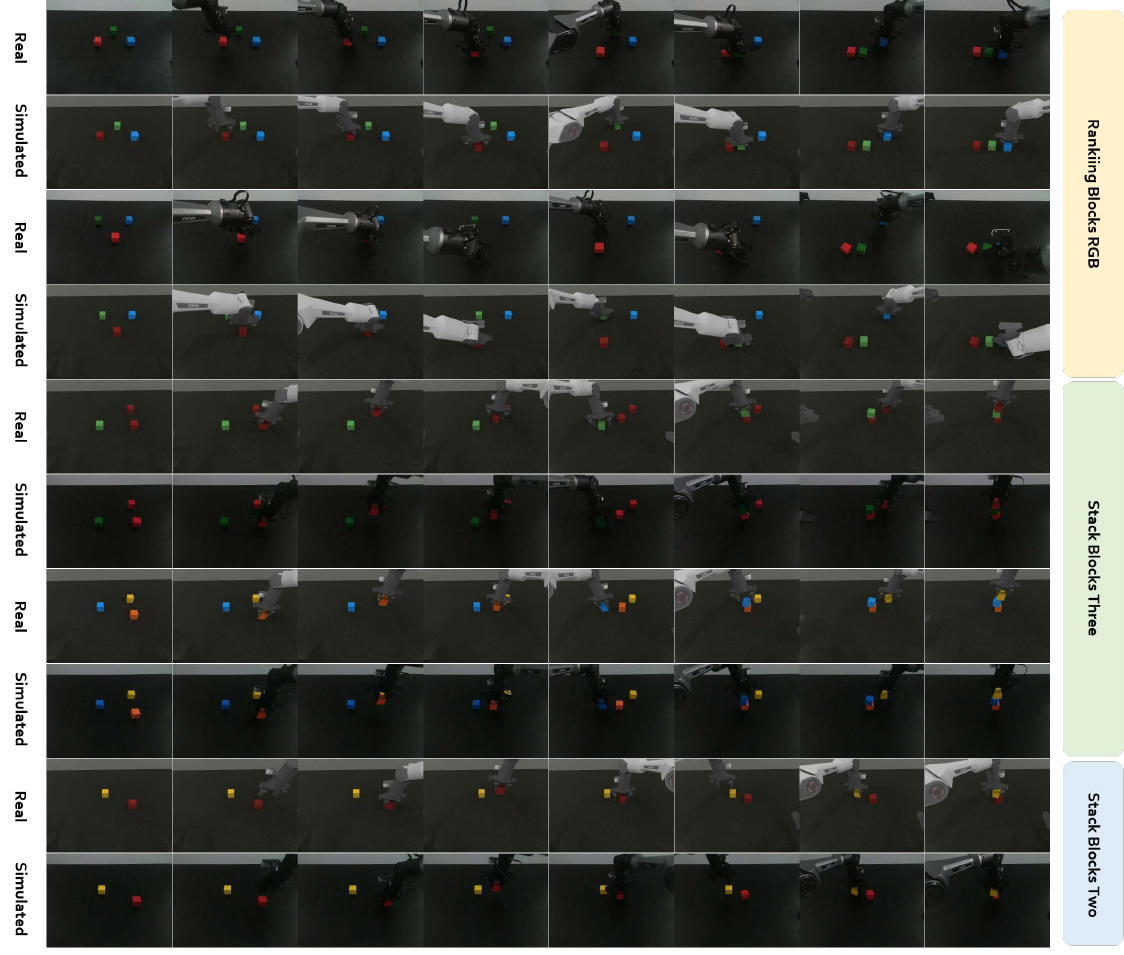}
\end{center}
\caption{Real2Sim alignment on additional tasks. From top to bottom: \textit{Ranking Blocks RGB}, \textit{Stack Blocks Three} and \textit{Stack Blocks Two}.}
\label{fig:visual_shake_other}
\label{fig:app_visual_other}
\end{figure}

As detailed in Sup.~\ref{app:sim2real_neural_simulation_details}, we performed exhaustive Real2Sim alignment. Here we illustrate the final alignment quality for the tasks \textit{Move Playing-Card Away}, \textit{Ranking Blocks RGB}, \textit{Stack Blocks Three} and \textit{Stack Blocks Two} in Fig.~\ref{fig:app_visual_shake_bottle} and Fig.~\ref{fig:app_visual_other}.

\subsection{Visualization of Sim2Real Neural Simulation}
\begin{figure}[H]
    \centering
    \includegraphics[width=1.0\linewidth]{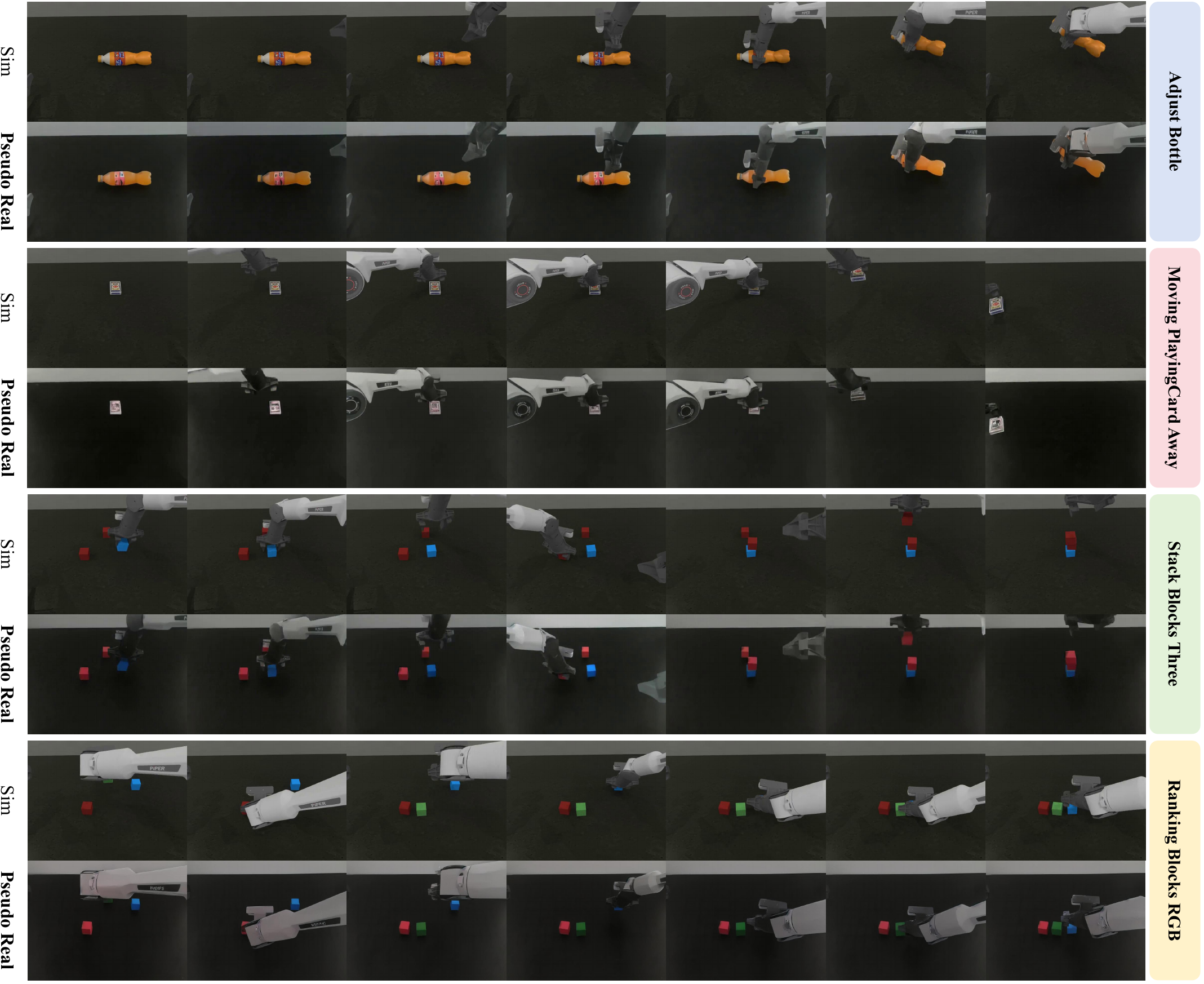}
    \caption{Sim2Real visualization on various tasks. From top to bottom: \textit{Adjust Bottle}, \textit{Moving PlayingCard Away}, \textit{Stack Blocks Three} and \textit{Ranking Blocks RGB}.}
    \label{fig:more_sim2real_viz}
\end{figure}

To dynamically demonstrate the effectiveness of our approach in sim-to-real transfer, we further present a visual comparison between the pseudo-realistic videos generated by our Neural Simulator and the initial simulation videos, as shown in Fig.~\ref{fig:more_sim2real_viz}. In addition to the tasks included in the main text—\textit{Adjust Bottle}, \textit{Moving PlayingCard Away}, and \textit{Ranking Blocks RGB}—we also consider the \textit{Stack Blocks Three} task. The results indicate that our method consistently maintains strong temporal coherence and perceptual realism throughout the video sequences.

\clearpage

\end{document}